\documentclass[sigconf]{acmart}

\usepackage{booktabs} 
\usepackage{algorithm}
\usepackage{algorithmic}
\usepackage{epsfig}
\usepackage{color}
\usepackage{flushend}
\usepackage{amsmath}
\usepackage{graphicx,subfigure}
\usepackage{multirow}
\usepackage{makecell}
\usepackage{CJKutf8}
\usepackage{balance}

\copyrightyear{2019}
\acmYear{2019}
\setcopyright{iw3c2w3}
\acmConference[WWW '19]{Proceedings of the 2019 World Wide Web Conference}{May 13--17,
	2019}{San Francisco, CA, USA}
\acmBooktitle{Proceedings of the 2019 World Wide Web Conference (WWW '19), May 13--17, 2019,
	San Francisco, CA, USA}
\acmPrice{}
\acmDOI{10.1145/3308558.3313743}
\acmISBN{978-1-4503-6674-8/19/05}

\begin{document}
\begin{CJK*}{UTF8}{gbsn}
	
\title{Neural Chinese Named Entity Recognition via \\ CNN-LSTM-CRF and Joint Training with Word Segmentation}

\author{Fangzhao Wu}
\affiliation{
	\institution{Microsoft Research Asia}
	\city{Beijing}
	\state{China}
}
\email{wufangzhao@gmail.com}

\author{Junxin Liu}
\affiliation{
	\institution{Tsinghua University}
	\city{Beijing}
	\state{China}
}
\email{ljx16@mails.tsinghua.edu.cn}

\author{Chuhan Wu}
\affiliation{
	\institution{Tsinghua University}
	\city{Beijing}
	\state{China}
}
\email{wuch15@mails.tsinghua.edu.cn}

\author{Yongfeng Huang}
\affiliation{
	\institution{Tsinghua University}
	\city{Beijing}
	\state{China}
}
\email{yfhuang@tsinghua.edu.cn}

\author{Xing Xie}
\affiliation{
	\institution{Microsoft Research Asia}
	\city{Beijing}
	\state{China}
}
\email{xingx@microsoft.com}

\begin{abstract}
Chinese named entity recognition (CNER) is an important task in Chinese natural language processing field.
However, CNER is very challenging since Chinese entity names are highly context-dependent.
In addition, Chinese texts lack delimiters to separate words, making it difficult to identify the boundary of entities.
Besides, the training data for CNER in many domains is usually insufficient, and annotating enough training data for CNER is very expensive and time-consuming.   
In this paper, we propose a neural approach for CNER.
First, we introduce a CNN-LSTM-CRF neural architecture to capture both local and long-distance contexts for CNER.
Second, we propose a unified framework to jointly train CNER and word segmentation models in order to enhance the ability of CNER model in identifying entity boundaries.
Third, we introduce an automatic method to generate pseudo labeled samples from existing labeled data which can enrich the training data.
Experiments on two benchmark datasets show that our approach can effectively improve the performance of Chinese named entity recognition, especially when training data is insufficient.
\end{abstract}

%
%



\maketitle

\section{Introduction}

The goal of named entity recognition (NER) is to identify entity names from texts and classify their types into different categories such as person, location and so on~\cite{dos2015boosting,misawa2017character}.
For example, given a sentence ``Bill Gates is the founder of Microsoft'', a NER tool can recognize that ``Bill Gates'' is a named entity of person and ``Microsoft'' is a named entity of organization.
NER is an important task in natural language processing field~\cite{MaH16,peng2017multi}, and an essential step for many downstream applications such as entity linking~\cite{han2011collective}, relation extraction~\cite{lin2016neural} and question answering~\cite{dong2015question}.

NER has been studied for many years and various methods have been proposed~\cite{LampleBSKD16,MaH16}.
Most of these methods are designed for NER of English texts.
For example, Settles~\shortcite{settles2004biomedical} proposed to use conditional random fields (CRF) for English medical text NER and incorporated both orthographic features and semantic features.
Huang et al.~\shortcite{huang2015bidirectional} proposed a LSTM-CRF neural model for English NER.
They combined spelling features (e.g., whether a word starts with a capital letter), context features (unigrams and bigrams), and word embeddings to build features for words.
Chiu and Nichols~\shortcite{ChiuN16} proposed to learn word- and character-level features using LSTM and CNN networks for English NER.
In their method, the word features include word embeddings, character embeddings learned from characters using CNN network, and capitalization features.

Compared with NER of English texts, Chinese NER is more difficult~\cite{levow2006third,dong2017multichannel}.
First, Chinese texts lack strong indications of entity names existing in English texts such as capitalization~\cite{peng2015named}.
Second, Chinese entity names are highly context-dependent.
Almost every Chinese character and word can be used in Chinese entity names.
The same word can be used as entity names or non-entity words in different contexts.
For example, in the sentence ``文献是朝阳区区长'', ``文献'' is a named entity of person.
However, in most cases the word ``文献'' is used as a non-entity word in Chinese texts with the meaning of ``literature''.
In addition, the same word can be used as names of different kind entities.
For instance, the word ``阿里'' can be the name of person entity (e.g., ``拳王阿里是个传奇''), location entity (``西藏阿里美不胜收''), and organization entity (``杭州阿里吸引了很多人才'').
Third, different from English texts, there is no delimiter such as whitespace to separate words in Chinese texts, making it more difficult to identify entity boundaries~\cite{LuoY16}.
For example, a Chinese NLP tool may segment the sentence ``习近平常与特朗普通电话'' into ``习近/平常/与/特朗/普通/电话'' and predicts that ``特朗'' is a person name, while in fact ``习近平'' and ``特朗普'' are correct named entities of person.
Besides, annotating sufficient data for training CNER models is very expensive and time-consuming, and the training data in many domains is limited.
Thus, CNER is quite challenging.

In order to handle these challenges, in this paper we propose a neural approach for Chinese named entity recognition.
First, in our approach we propose a CNN-LSTM-CRF architecture to fully capture both local and long-distance contexts for CNER.
Our model contains three layers, i.e., a convolutional
neural network (CNN)~\cite{lecun1989backpropagation} layer to learn contextual character representations from local contexts, a bidirectional long-short term memory (Bi-LSTM)~\cite{hochreiter1997long} layer to learn contextual character representations from long-distance contexts, and a CRF layer to jointly decode character labels.
In addition, we propose a unified framework to jointly train CNER and Chinese word segmentation (CWS) models in order to improve the ability of CNER model in predicting entity boundaries.
In our framework, the CWS model shares the same character embeddings and CNN network with the CNER model, and has an independent CRF layer for label decoding.
Besides, we propose an automatic method to generate pseudo labeled samples based on existing labeled data by randomly replacing entity names in labeled sentences with other entity names of the same concept.
These pseudo labeled samples are usually correct in grammar and smooth in semantics, and can enrich the training data and improve the generalization ability of CNER model.
Extensive experiments are conducted on two benchmark datasets.
The experimental results show that our approach can significantly improve the performance of Chinese named entity recognition and outperform many baseline methods.

\section{Related Work}\label{sec:RelatedWork}

Chinese named entity recognition is usually modeled as a character-level sequence labeling problem~\cite{gao2005chinese,liu2010chinese}, since Chinese sentence is a string of characters and there is no explicit delimiter like whitespace to separate characters into words.
Statistical sequence modeling methods such as CRF~\cite{lafferty2001conditional} are widely used in Chinese NER~\cite{LuoY16,peng2015named}.
A core step in these methods is building feature representation for each character in sentence.
Traditionally, these character features are manually designed~\cite{chen2006chinese,yu2008chinese}.
For example, character unigrams, bigrams and clusters are used as character features in~\cite{chen2006chinese}.
Zhang et al.~\shortcite{zhang2008fusion} proposed to build character features using surrounding characters and whether target character is in external dictionaries.
Yu et al.~\shortcite{yu2008chinese} proposed to combine character-level features, word-level features, POS tagging features and dictionary features for CNER.
These handcrafted features require a large amount of domain knowledge to design and many of them rely on external resources such as gazetteers, which may not exist in many domains.
Moreover, handcrafted features such as character n-grams usually cannot capture global context information.
Different from above methods, our approach can learn character feature representations from data without feature engineering, and can incorporate both local and long-distance context information.

In recent years, neural network methods have been applied to English NER~\cite{huang2015bidirectional,LampleBSKD16}.
Most of these methods are based on LSTM-CRF architecture.
For example, Lample et al.~\shortcite{LampleBSKD16} used Bi-LSTM to learn hidden representations of words and used CRF to do label decoding. 
These methods for English NER are usually based on word-level input, and cannot be directly applied to Chinese NER which is a character-level sequence labeling task.
CNN is exploited in~\cite{ChiuN16} and~\cite{MaH16} to learn word representations from characters for English NER.
These word representations are combined with word embeddings and/or other features to build word features.
However, there is no explicit word boundaries in Chinese texts.
Thus, these CNN incorporated methods for English NER are not suitable for Chinese NER.
Different from these methods, in our approach CNN is used to learn representations of characters rather than words.

Recently, neural network methods are also applied to Chinese NER~\cite{dong2016character,he2017unified,dong2017multichannel}.
These methods are based on LSTM-CRF framework, where LSTM is used to learn hidden representations of characters and CRF is used for joint labeling decoding.
Many of these methods also incorporate handcrafted features, such as radical features~\cite{dong2016character} and position features~\cite{he2017unified}, to build character representations.
Different from these methods, our approach can train CNER model in an end-to-end manner and does not need any manual feature engineering.
In addition, our approach is based on the CNN-LSTM-CRF architecture to fully capture both local and long-distance contexts.

Jointly training NER model with related tasks has the potential to improve the performance of NER~\cite{luo2015joint,peng2017multi}.
For example, Luo et al.~\shortcite{luo2015joint} proposed to jointly train NER and entity linking models, and achieved state-of-the-art performance on English NER.
Since Chinese texts lack word delimiters and identifying entity boundaries is very challenging in CNER, we propose to jointly train CNER model with word segmentation to improve the ability of CNER model in identifying entity boundaries. 
Similar with our approach, Peng et al.~\shortcite{peng2017multi} proposed to jointly train CNER and CWS models in domain adaptation scenario.
In their method, CNER and CWS share the same character representation model and have different label decoding models.
In our approach we regard word segmentation as an auxiliary task and use it to learn word boundary aware character representations for CNER.
Experiments show that our approach is more effective.
In addition, we propose to automatically generate pseudo labeled samples from labeled data, which to our best knowledge has not been explored in existing CNER methods.

\section{Our Approach}\label{sec:method}

In this section we introduce our approach for CNER in detail.
First, we present the CNN-LSTM-CRF neural architecture for CNER.
Next, we introduce a unified framework to jointly train CNER and word segmentation models.
Finally, we introduce an automatic method to generate pseudo labeled samples from existing labeled data.  

\subsection{CNN-LSTM-CRF Architecture for CNER}

\begin{figure}[tp]
	\centering
	\includegraphics[width=0.48\textwidth]{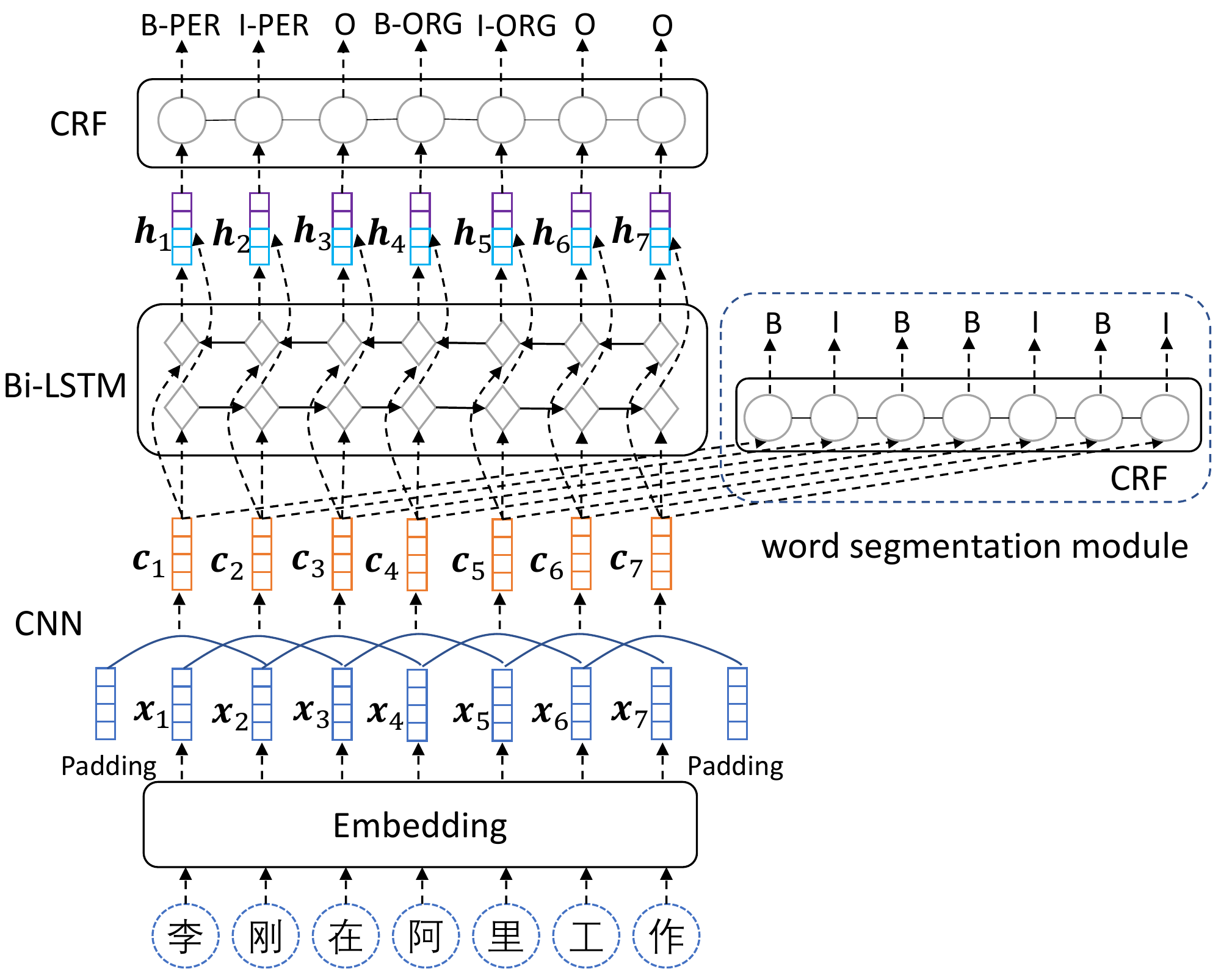}
	\caption{The framework of our approach.}\label{fig.model}
\end{figure}

Our proposed CNN-LSTM-CRF neural architecture for Chinese named entity recognition is illustrated in Fig.~\ref{fig.model}. 
Next we introduce each layer from bottom to top in detail.

The first layer is character embedding, which aims to convert a sentence from a sequence of characters into a sequence of dense vectors.
In this layer, an embedding matrix $\mathbf{E}\in \mathcal{R}^{D\times V}$ is used to map each character into a dense vector, where $D$ is embedding dimension and $V$ is vocabulary size.
Denote an input sentence as $s = [w_1, w_2, ..., w_N]$, where $N$ is sentence length and $w_i \in \mathcal{R}^{V}$ is the one-hot representation of the $i_{th}$ character.
The output of this layer is a sequence of vectors $[\mathbf{x}_1, \mathbf{x}_2, ..., \mathbf{x}_N]$, where $\mathbf{x}_i = \mathbf{E}w_i \in \mathcal{R}^{D}$.

The second layer is a CNN network.
CNN has been widely used in computer vision field to extract local information of images~\cite{lecun2015deep}.
The local context of sentences is also very important for Chinese named entity recognition.
For example, the word ``阿里'' is a location entity in ``西藏阿里'', while is a organization entity in ``杭州阿里''.
Motivated by these observations, we propose to use CNN to capture the local context information for CNER.
Denote $\mathbf{w} \in \mathcal{R}^{KD}$ as a filter in CNN where $K$ is the window size, then the contextual representation of the $i_{th}$ character learned by this filter is:
\begin{equation}\label{eq.filter}
	\begin{split}
		c_i = f(\mathbf{w}^T \times \mathbf{x}_{\lfloor i-\frac{K-1}{2}\rfloor:\lfloor i+\frac{K-1}{2}\rfloor}),
	\end{split}
\end{equation} 
where $\mathbf{x}_{\lfloor i-\frac{K-1}{2}\rfloor:\lfloor i+\frac{K-1}{2}\rfloor}$ represents the concatenation of the embeddings of characters from $\lfloor i-\frac{K-1}{2}\rfloor$ to $\lfloor i+\frac{K-1}{2}\rfloor$, and $f$ is the activation function. 
Here we select ReLU~\cite{lecun2015deep} as the activation function.
In this layer, we use multiple filters with different window sizes (ranging from 2 to 5) to learn contextual character representations.
Denote the filter number as $M$.
The contextual representation of the $i_{th}$ character (denoted as $\mathbf{c}_i$) is the concatenation of the outputs of all filters at this position.
The output of the CNN layer is $\mathbf{c} = [\mathbf{c}_1,\mathbf{c}_2,...,\mathbf{c}_N]$, where $\mathbf{c}_i \in \mathcal{R}^{M}$.

The third layer is a Bi-LSTM network.
LSTM is a special type of recurrent neural networks (RNN) which can capture long-distance sequence information and is powerful in modeling sequential data.
LSTM can generate hidden states for tokens in a sequence using all previous contexts, which is beneficial for Chinese named entity recognition.
For example, in sentence ``阿里是中国非常著名的IT企业'' (Alibaba is a famous Chinese IT company), although ``IT企业'' has a long distance with ``阿里'',  it is very important for inferring that ``阿里'' is an organization entity. 
In addition, both left and right contexts can be useful for recognizing named entities.
Thus, we use bidirectional LSTM (Bi-LSTM) to learn hidden representations of characters from global contexts.
Denote $\overrightarrow{lstm}$ as the LSTM scanning text from left to right, and $\overleftarrow{lstm}$ as the LSTM scanning text from right to left, then the hidden representations learned by these LSTMs can be represented as $[\overrightarrow{\mathbf{h}_1},\overrightarrow{\mathbf{h}_2},...,\overrightarrow{\mathbf{h}_N}]=\overrightarrow{lstm}([\mathbf{c}_1,\mathbf{c}_2,...,\mathbf{c}_N])$ and $[\overleftarrow{\mathbf{h}_1},\overleftarrow{\mathbf{h}_2},...,\overleftarrow{\mathbf{h}_N}]=\overleftarrow{lstm}([\mathbf{c}_1,\mathbf{c}_2,...,\mathbf{c}_N])$.
The hidden representation of the $i_{th}$ character is a concatenation of $\overrightarrow{\mathbf{h}_i}$ and $\overleftarrow{\mathbf{h}_i}$, i.e., $\mathbf{h}_i = [\overrightarrow{\mathbf{h}_i}, \overleftarrow{\mathbf{h}_i}]$.
The output of the Bi-LSTM layer is $\mathbf{h} = [\mathbf{h}_1,\mathbf{h}_2,...,\mathbf{h}_N]$, where $\mathbf{h}_i \in \mathcal{R}^{2S}$ and $S$ is the dimension of hidden states in LSTM.

The fourth layer in our model is conditional random field (CRF).
In NER, neighboring labels usually have strong dependencies~\cite{MaH16}.
For example, \textit{I-PER} label usually follows \textit{B-PER} and \textit{I-PER}, but it cannot follow \textit{B-ORG} or \textit{I-ORG}.
Thus, it is beneficial to jointly decode the labels of characters in a sequence rather than decode them independently~\cite{collobert2011natural}.
In this paper, we use first-order linear chain CRF~\cite{lafferty2001conditional} to jointly decode the labels of characters.

Denote $\mathbf{y} = [y_1, y_2,...,y_N]$ as the label sequence of sentence $s$, where $y_i \in \mathcal{R}^{L}$ is the one-hot representation of the $i_{th}$ character's label, and $L$ is the number of labels.
The input of the CRF layer is the hidden representations of characters generated by the Bi-LSTM layer, i.e., $\mathbf{h} = [\mathbf{h}_1, \mathbf{h}_2,...,\mathbf{h}_N]$, and the output of the CRF layer is the label sequence $\mathbf{y}$.
CRF is a probabilistic model.
The conditional probability of label sequence $\mathbf{y}$ given input $\mathbf{h}$ is computed as follows:
\begin{equation}\label{eq.posterior}
	\begin{split}
		p(\mathbf{y}|\mathbf{h}; \mathbf{\theta}) = \frac{ \prod\limits_{i=1}^{N}\psi(\mathbf{h}_i, y_i, y_{i-1})}{\sum\limits_{\mathbf{y}^\prime \in \mathcal{Y}(s)} \prod\limits_{i=1}^{N}\psi(\mathbf{h}_i, y_i^\prime, y_{i-1}^\prime)},
	\end{split}
\end{equation} 
where $\mathcal{Y}(s)$ is the set of all possible label sequences of sentence $s$, $\mathbf{\theta}$ is the parameter set, and $\psi(\mathbf{h}_i, y_i, y_{i-1})$ is the potential function.
In our approach, the potential function is:
\begin{equation}\label{eq.potential}
	\begin{split}
		\psi(\mathbf{h}_i, y_i, y_{i-1}) = \exp(y_i^T\mathbf{W}^T\mathbf{h}_i+y_{i-1}^T\mathbf{T}y_i),
	\end{split}
\end{equation}  
where $\mathbf{W} \in \mathcal{R}^{2S\times L}$ and $\mathbf{T} \in \mathcal{R}^{L\times L}$ are parameters of the CRF layer, and $\mathbf{\theta} = \{\mathbf{W}, \mathbf{T}\}$ in Eq.~(\ref{eq.posterior}). 

The loss function of our approach is the negative log-likelihood over all training samples, which can be formulated as follows:
\begin{equation}\label{eq.loss}
	\begin{split}
		\mathcal{L}_{NER} = - \sum_{s\in \mathcal{S}} \log(p(\mathbf{y}_s|\mathbf{h}_s; \mathbf{\theta})),
	\end{split}
\end{equation} 
where $\mathcal{S}$ is the set of sentences in training data, $\mathbf{h}_s$ and $\mathbf{y}_s$ are the hidden representations and label sequence of sentence $s$.

\subsection{Joint Training with Word Segmentation}

The NER task can be regarded as a combination of two sub-tasks: extracting entity names from texts (i.e., identifying the boundaries of entity names) and classifying their types.
Identifying the boundaries of Chinese entity names is quite challenging, since there is no explicit word delimiters in Chinese texts.
For example, many existing Chinese NLP tools such as LTP predict that ``特朗'' is an entity name in the sentence ``习近平常与特朗普通电话'', while the correct entity name is ``特朗普''.

In Chinese natural language processing field, the goal of Chinese word segmentation (CWS) is segmenting Chinese texts into words (in other words, predicting the boundaries of words in texts).
Thus, CWS is highly related to CNER and has the potential to help CNER predict entity boundaries more accurately.
Motivated by this observation, we propose to enhance the ability of CNER model in identifying entity boundaries by jointly training CNER and CWS models.
Similar with CNER, CWS is also usually modeled as a character-level sequence labeling problem~\cite{gao2005chinese,chen2015long,liu2018neural}, and CRF is widely used in CWS methods.
Our unified framework for jointly training CNER with CWS is shown in Fig.~\ref{fig.model}.
In our framework the CNER and CWS models share the same character embeddings and CNN network.
In this way, the useful information in word segmentation can be encoded to learn word boundary aware contextual character representations, which is useful for predicting entity boundaries.
The loss function of the CWS module is:
\begin{equation}\label{eq.loss_CWS}
	\begin{split}
		\mathcal{L}_{CWS} = - \sum_{s\in \mathcal{S}} \log(p(\mathbf{y}_s^{seg}|\mathbf{c}_s; \mathbf{\theta}^{seg})),
	\end{split}
\end{equation} 
where $\mathbf{y}_s^{seg}$ is the label sequence of sentence $s$ for word segmentation, and $\mathbf{\theta}^{seg}$ is the parameter set of CWS model.
$\mathbf{c}_s$ is the hidden character representations of sentence $s$ output by the CNN network.

The final objective function of our approach is the combination of the CNER loss and the CWS loss as follows:
\begin{equation}\label{eq.loss_all}
	\begin{split}
		\mathcal{L} = (1-\lambda)\mathcal{L}_{NER} + \lambda\mathcal{L}_{CWS},
	\end{split}
\end{equation} 
where $\lambda \in [0,1)$ is a coefficient to control the relative importance of the CWS loss in the overall loss.

\subsection{Pseudo Labeled Data Generation}

The labeled data for Chinese named entity recognition is usually scarce in many domains.
Manually annotating sufficient samples for CNER is time-consuming and expensive.
Thus, an automatic method to generate labeled samples for CNER will be very useful.
In this paper we propose to automatically generate pseudo labeled samples based on existing labeled data.
Our method is motivated by the observation that if an entity name in a sentence is replaced by another entity name with the same concept, then the new sentence is usually correct in grammar and semantics.
For example, a Chinese sentence may be ``李刚在阿里工作", where ``李刚" is a person entity and ``阿里" is a company entity.
If ``王小超" is another person name and ``谷歌" is another company name, then we can obtain a new sentence by replacing ``李刚" with ``王小超" and ``阿里" with ``谷歌", i.e., ``王小超在谷歌工作".
If we have the NER labels of the original sentence, then we can easily infer the NER labels of the pseudo sentence.
For instance, if the NER labels of the original sentence in aforementioned example are ``B-PER/I-PER/O/B-ORG/I-ORG/O/O'', then the NER labels of the pseudo sentence are ``B-PER/I-PER/I-PER/O/B-ORG/I-ORG/O/O''.
In addition, if we have CWS labels of the original sentence, then we can also automatically obtain CWS labels of the pseudo sentence, since entity names are usually regarded as independent words.
For instance, in aforementioned example, if the CWS labels of the original sentence are ``B/I/B/B/I/B/B/I", then the CWS labels of the pseudo sentence are ``B/I/I/B/B/I/B/B/I".

In our method, given a set of labeled samples for CNER, we first extract all the entity names (denoted as $\mathit{EN}$) from them.
Then we randomly select a labeled sentence, and replace each entity in this sentence with an entity randomly sampled from $\mathit{EN}$ which has the same concept.
In this way a pseudo sentence is generated and its NER labels (as well as CWS labels) can be automatically inferred.
This step is repeated multiple times until a predefined number of pseudo labeled samples are generated.

\section{Experiments} \label{sec:Experiments}

\subsection{Datasets and Experimental Settings}

Two benchmark datasets for Chinese named entity recognition are used in our experiments.
The first one is the MSRA corpus released by the third SIGHAN Chinese language processing bakeoff\footnote{{http://sighan.cs.uchicago.edu/bakeoff2006/download.html}} (denoted as \emph{Bakeoff-3}).
It contains 46,364 sentences for training and 4,365 sentences for test.
The second dataset is the MSRA corpus released by the fourth SIGHAN Chinese language processing bakeoff\footnote{{https://www.aclweb.org/mirror/ijcnlp08/sighan6/chinesebakeoff.htm}} (denoted as \emph{Bakeoff-4}).
This dataset contains 23,181 training sentences and 4,636 test sentences.
Both datasets contain three types of named entities, i.e., \textit{person} (denoted as \emph{PER}), \textit{location} (denoted as \emph{LOC}) and \textit{organization} (denoted as \emph{ORG}).
The statistics of these datasets are summarized in Table~\ref{table.dataset}.

\begin{table}[!htb]
	\caption{The statistics of datasets. \emph{\#Sen} and \emph{\#Ent} represent the numbers of sentences and entities.}\label{table.dataset}
	\centering
	\resizebox{0.45\textwidth}{!}{
		\begin{tabular}{@{\extracolsep{0pt}} c| c| c |c |c |c }
			\Xhline{1pt}
			\multirow{2}{*}{Dataset} &
			\multicolumn{2}{c|}{\emph{Training}} &
			\multicolumn{3}{c}{\emph{Test}} \\
			\Xcline{2-6}{0.5pt}
			& \emph{\#Sen} & \emph{\#Ent}  & \emph{\#Sen}  & \emph{\#Ent}  &\emph{OOV Rate}\\
			\Xhline{0.9pt}
			\emph{Bakeoff-3}&46,364&74,528&4,365&6,190&28.16\%\\
			\hline
			\emph{Bakeoff-4}&23,181&37,811&4,636&7,707&21.42\%\\
			\Xhline{1pt}
		\end{tabular}
	}
	
\end{table}

\begin{table*}[!htb]
	\caption{The performance of different methods on the \emph{Bakeoff-3} dataset. $P$, $R$ and $F$ represent precision, recall and Fscore respectively. $R_{oov}$ stands for the recall of out-of-vocabulary entities.}\label{table.performance1}
	\centering
	\resizebox{0.9\textwidth}{!}{
		\begin{tabular}{@{\extracolsep{0pt}} c| c| c| c| c| c| c| c| c| c| c| c| c}
			\Xhline{1pt}
			\multirow{2}{*}{} &
			\multicolumn{4}{c|}{5\%} & \multicolumn{4}{c|}{25\%} & \multicolumn{4}{c}{100\%} \\
			\Xcline{2-13}{0.5pt}
			& $P$ & $R$  & $F$ & $R_{oov}$  & $P$ & $R$  & $F$ & $R_{oov}$ & $P$ & $R$  & $F$  & $R_{oov}$  \\
			\Xhline{0.9pt}
			\emph{LSTM+Softmax}     & 56.14 & 49.83 & 52.68 & 33.03 & 73.02 & 66.94 & 69.84 & 49.00 & 82.76 &	78.52 & 80.59 & 60.57 \\
			\hline
			\emph{LSTM+CRF}         & 61.00 & 54.32 & 57.37 & 37.09 & 77.75 & 70.42 & 73.90 & 53.05 & 87.30 &	82.93 & 85.06 & 68.06\\
			\hline
			\hline
			\emph{MDA}              & 65.12 & 57.90 & 61.27 & 42.74 & 79.71 & 73.16 & 76.29 & 58.34 & 87.78 &	82.43 & 85.02 & 67.15\\
			\hline
			\emph{LSTM+CRF+Pseudo}  & 61.21 & 57.25 & 59.07 & 40.22 & 79.57 & 75.80 & 77.63 & 60.45 & 88.13 &	85.74 & 86.92 & 73.19\\
			\hline
			\hline
			\emph{CNN+LSTM+Softmax} & 62.72 & 56.96 & 59.62 & 39.75 & 77.83 & 74.54 & 76.14 & 57.15 & 87.60 &	83.50 & 85.50 & 67.41 \\
			\hline
			\emph{CLC}              & 65.58 & 61.60 & 63.45 & 43.04 & 81.22 & 76.20 & 78.59 & 59.23 & 90.39 &	86.43 & 88.37 & 71.54 \\
			\hline
			\emph{CLC+Joint}        & 70.70 & 62.93 & 66.56 & 45.78 & 81.20 & 78.32 & 79.72 & 63.13 & 90.29 &	86.98 & 88.60 & 73.61 \\
			\hline
			\emph{CLC+Pseudo}       & 69.86 & 64.13 & 66.84 & 49.17 & 82.69 & 79.93 & 81.27 & 65.38 & 89.59 &	88.46 & 89.02 & 76.99 \\
			\hline
			\emph{CLC+Joint+Pseudo} & 70.31 & 66.51 & 68.33 & 49.30 & 83.57 & 81.62 & 82.57 & 66.86 & 89.48 &	89.36 & 89.42 & 77.93 \\
			\Xhline{1pt}
		\end{tabular}
	}
\end{table*}

\begin{table*}[!htb]
	\caption{The performance of different methods on the \emph{Bakeoff-4} dataset.}\label{table.performance2}
	\centering
	\resizebox{0.9\textwidth}{!}{
		\begin{tabular}{@{\extracolsep{0pt}} c| c| c| c|c| c| c|c|c| c|c| c| c  }
			\Xhline{1pt}
			\multirow{2}{*}{} &
			\multicolumn{4}{c|}{5\%} & \multicolumn{4}{c|}{25\%} & \multicolumn{4}{c}{100\%} \\
			\Xcline{2-13}{0.5pt}
			& $P$ & $R$  & $F$ & $R_{oov}$  & $P$ & $R$  & $F$ & $R_{oov}$ & $P$ & $R$  & $F$  & $R_{oov}$  \\
			\Xhline{0.9pt}
			\emph{LSTM+Softmax}     & 51.78 & 42.88 & 46.74 & 30.24 & 69.32 & 67.16 & 68.18 & 49.62 & 83.39 & 80.31 & 81.81 & 63.08 \\
			\hline
			\emph{LSTM+CRF}         & 54.18 & 47.53 & 50.48 & 30.64 & 74.90 & 71.64 & 73.21 & 56.07 & 87.10 & 85.07 & 86.07 & 69.96 \\
			\hline
			\hline
			\emph{MDA}              & 56.66 & 49.75 & 52.90 & 34.93 & 76.29 & 73.91 & 75.05 & 59.09 & 87.56 & 85.01 & 86.26 & 70.77 \\
			\hline
			\emph{LSTM+CRF+Pseudo}  & 57.37 & 49.54 & 53.09 & 34.53 & 76.31 & 75.79 & 76.02 & 59.85 & 86.94 & 86.90 & 86.92 & 72.07 \\
			\hline
			\hline
			\emph{CNN+LSTM+Softmax} & 59.09 & 51.12 & 54.74 & 35.81 & 76.42 & 73.74 & 74.99 & 58.13 & 88.48 & 86.34 & 87.40 & 69.47 \\
			\hline
			\emph{CLC}              & 58.42 & 53.56 & 55.83 & 39.41 & 78.84 & 78.50 & 78.59 & 61.82 & 89.86 & 88.74 & 89.30 & 73.97 \\
			\hline
			\emph{CLC+Joint}        & 61.88 & 56.23 & 58.76 & 41.49 & 82.87 & 78.63 & 80.65 & 63.43 & 90.23 & 88.80 & 89.51 & 74.86 \\
			\hline
			\emph{CLC+Pseudo}       & 61.82 & 59.69 & 60.67 & 43.40 & 82.33 & 80.41 & 81.35 & 64.00 & 90.44 & 89.80 & 90.12 & 75.00 \\
			\hline
			\emph{CLC+Joint+Pseudo} & 64.21 & 61.58 & 62.78 & 47.58 & 83.16 & 81.93 & 82.52 & 66.88 & 90.09 & 90.28 & 90.18 & 77.13 \\
			\Xhline{1pt}
		\end{tabular}
	}
\end{table*}

Following~\cite{wang2017named}, we use \emph{BIO} tagging scheme for CNER, where \emph{B}, \emph{I} and \emph{O} respectively represent the beginning, inside and outside of entity name.
The labels \emph{B} and \emph{I} are further combined with entity types, such as ``B-PER" and ``I-LOC".
As for CWS, we use the \emph{BI} tagging scheme.
In the \emph{Bakeoff-3} dataset the annotation of word segmentation in training set is provided, while in the \emph{Bakeoff-4} dataset it is not provided.
Thus, we used LTP\footnote{https://www.ltp-cloud.com/intro/en/} tool to segment the sentences in the training set of the \emph{Bakeoff-4} dataset.
Note that the word segmentation information is not involved in the test data.

In our experiments, the embeddings of characters are pretrained on the Sogou News Corpus\footnote{http://www.sogou.com/labs/resource/ca.php} using the word2vec tool\footnote{https://code.google.com/archive/p/word2vec/}.
The dimension of character embedding is 200.
These embeddings are tuned during model training.
Since dropout method~\cite{srivastava2014dropout} is an effective way to mitigate overfitting, we apply it to the outputs of embedding, CNN and LSTM layers. 
Our approach is implemented using Tensorflow library, and RMSProp~\cite{dauphin2015equilibrated} is used as the optimization algorithm to learn model parameters.
Hyperparameters are selected according to cross-validation on training data.
In detail, the number of filters in CNN network is 400, the hidden state size of LSTM is 200, the dropout rate is 0.2, and the $\lambda$ in Eq.~(\ref{eq.loss_all}) is 0.4.
In each experiment, we randomly sample 10\% of the training data for validation and the remaining for model training.
Each experiment is repeated 10 times, and the average results are reported.

\subsection{Performance Evaluation}

First we compare the performance of our approach with different neural network based CNER methods, including: 1) \textit{LSTM+CRF}, using Bi-LSTM to learn character representations and CRF to decode labels~\cite{peng2017multi}; 2) \textit{LSTM+Softmax}, using Bi-LSTM to learn character representations and softmax for label decoding; 3) \textit{MDA}, multi-task domain adaptation method proposed by Peng et al.~\shortcite{peng2017multi}, which jointly trains CNER and CWS models via multi-task learning; 4) \textit{LSTM+CRF+Pseudo}, \textit{LSTM+CRF} trained on both original and pseudo labeled samples; 5) \textit{CLC}, our CNN-LSTM-CRF architecture for CNER; 6) \textit{CNN+LSTM+Softmax}, replacing the CRF layer in \textit{CLC} with a softmax layer; 7) \textit{CLC+Joint}, our unified framework for jointly training CNER and CWS models; 8) \textit{CLC+Pseudo}, \textit{CLC} trained on both original and pseudo labeled samples; 9) \textit{CLC+Joint+Pseudo}, \textit{CLC} with both joint training and pseudo labeled data.

We conducted experiments on different ratios (i.e., 5\%, 25\% and 100\%) of training data to test the performance of different methods with insufficient and sufficient labeled samples.
For those methods which involve pseudo labeled samples, the number of pseudo samples is the same with the real labeled samples.
The experimental results on the \emph{Bakeoff-3} and \emph{Bakeoff-4} datasets are shown in Tables~\ref{table.performance1} and \ref{table.performance2} respectively.
We have several findings from the results.

First, our CNN-LSTM-CRF architecture (\textit{CLC}) performs better than the popular \textit{LSTM-CRF} architecture~\cite{peng2017multi} on Chinese NER.
For example, on the \textit{Bakeoff-3} dataset, the improvement of \textit{CLC} over \textit{LSTM-CRF} in terms of Fscore is 6.08\% on 5\% training data and 3.31\% on 100\% training data.
These results validate that our CNN-LSTM-CRF architecture is more suitable for Chinese NER than LSTM-CRF.
This is because our model can capture both local and long-distance contexts of sentences to learn contextual character representations, which is useful for Chinese NER.

Second, joint label decoding via CRF is better than independent label decoding for Chinese NER .
For instance, \textit{CLC} performs much better than \textit{CNN+LSTM+Softmax}, and \textit{LSTM+CRF} can consistently outperform \textit{LSTM+Softmax}.
This is because there are strong dependencies between neighboring NER labels, and CRF can effectively capture these dependencies.
In addition, \textit{CNN+LSTM+Softmax} consistently outperforms \textit{LSTM+Softmax}, although they both use independent label decoding.
It further validates better character representations can be learned by combing CNN and LSTM to capture both local and long-distance contexts.

Third, by jointly training CNER and CWS models using our unified framework, our approach (i.e., \textit{CLC+Joint}) can achieve better performance, especially when training data is insufficient.
This is because there is inherent relatedness between CNER and CWS, since CNER model needs to identify entity boundaries and the goal of CWS is to identify word boundaries.
Experimental results show that our unified framework is effective in exploiting the inherent relatedness between CNER and CWS, and can improve the ability of CNER model in identifying entity boundaries via jointly training CNER and CWS models.
\textit{MDA} method~\cite{peng2017multi} also combines CNER and CWS.
Our \textit{CLC+Joint} approach can consistently outperform \textit{MDA}.
There are two major differences between our approach and \textit{MDA}.
First, our approach is based on the CNN-LSTM-CRF architecture while \textit{MDA} is based on LSTM-CRF.
Second, in \textit{MDA} method CNER and CWS are regarded as equally important and share the same character representations, while in our approach CWS is regarded as an axillary task and used to learn word boundary aware character representations from local contexts.
These results indicate our approach is more suitable to exploit the useful information in word segmentation for CNER.

Fourth, the automatically generated pseudo labeled samples can effectively improve the performance of our approach and baseline methods.
For example, \textit{CLC+Pseudo} significantly outperforms \textit{CLC}, and \textit{LSTM+CRF+Pseudo} consistently outperforms \textit{LSTM+CRF}, especially in terms of OOV recall.
Besides, the performance improvement becomes bigger when training data is insufficient.
These results validate that the pseudo labeled samples automatically generated from labeled data can provide useful information for training CNER model.
They can improve the generalization ability of CNER model by encouraging it to capture the contextual patterns of entity names, which is important for recognizing OOV entities.
Moreover, accoding to Tables~\ref{table.performance1} and \ref{table.performance2}, after incorporating both pseudo labeled samples and joint training with CWS, the performance of our approach (\textit{CLC+Joint+Pseudo}) can be further improved.

\subsection{Influence of Pseudo Labeled Samples} 

\begin{figure}[tp]
	\centering
	\includegraphics[width=0.35\textwidth]{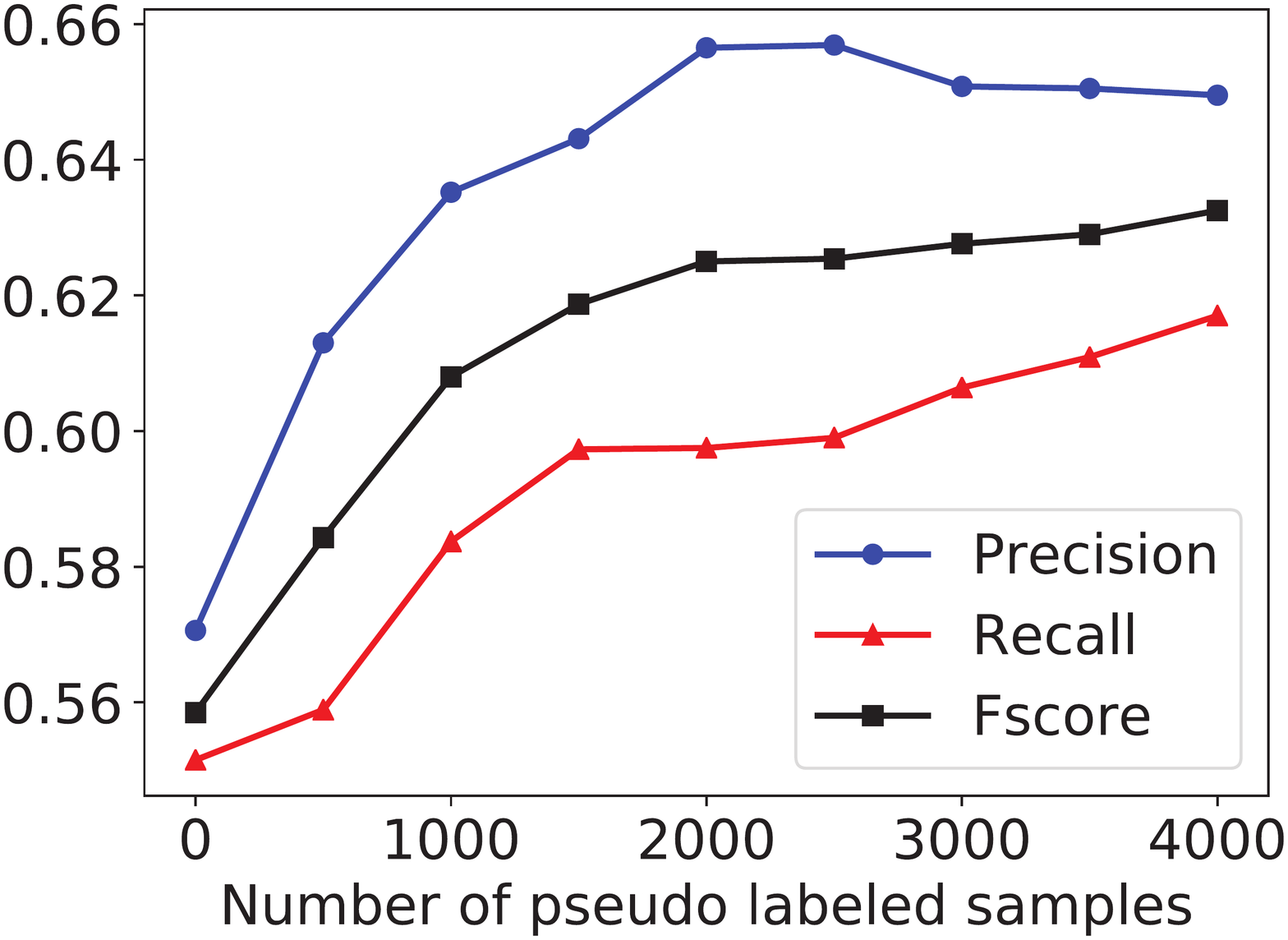}
	\caption{The influence of pseudo labeled data size.}\label{fig.pseudo_num}
\end{figure}

Next we explore the influence of the number of automatically generated pseudo labeled samples on the performance of our approach.
We conducted experiments on both datasets and the results on the \textit{Bakeoff-4} dataset with 5\% training data are shown in Figure~\ref{fig.pseudo_num}.
The results on the \textit{Bakeoff-3} dataset show similar patterns.

According to Figure~\ref{fig.pseudo_num}, as the number of pseudo labeled samples increases, the recall and Fscore of our approach consistently improve, while the precision first increases and slightly declines afterwards.
These results show that the pseudo labeled samples automatically generated from labeled data can provide useful information for training CNER model and can improve its ability in capturing contextual patterns of entity names, which is beneficial for identifying OOV entities.
Thus, the recall of our approach continuously improves when more pseudo labeled samples are incorporated.
However, when too many pseudo labeled samples are used, the CNER model may overfit these patterns, resulting in the slight decline of precision.

\subsection{Influence of Hyperparameters} 

\begin{figure}[tp]
	\centering
	\includegraphics[width=0.35\textwidth]{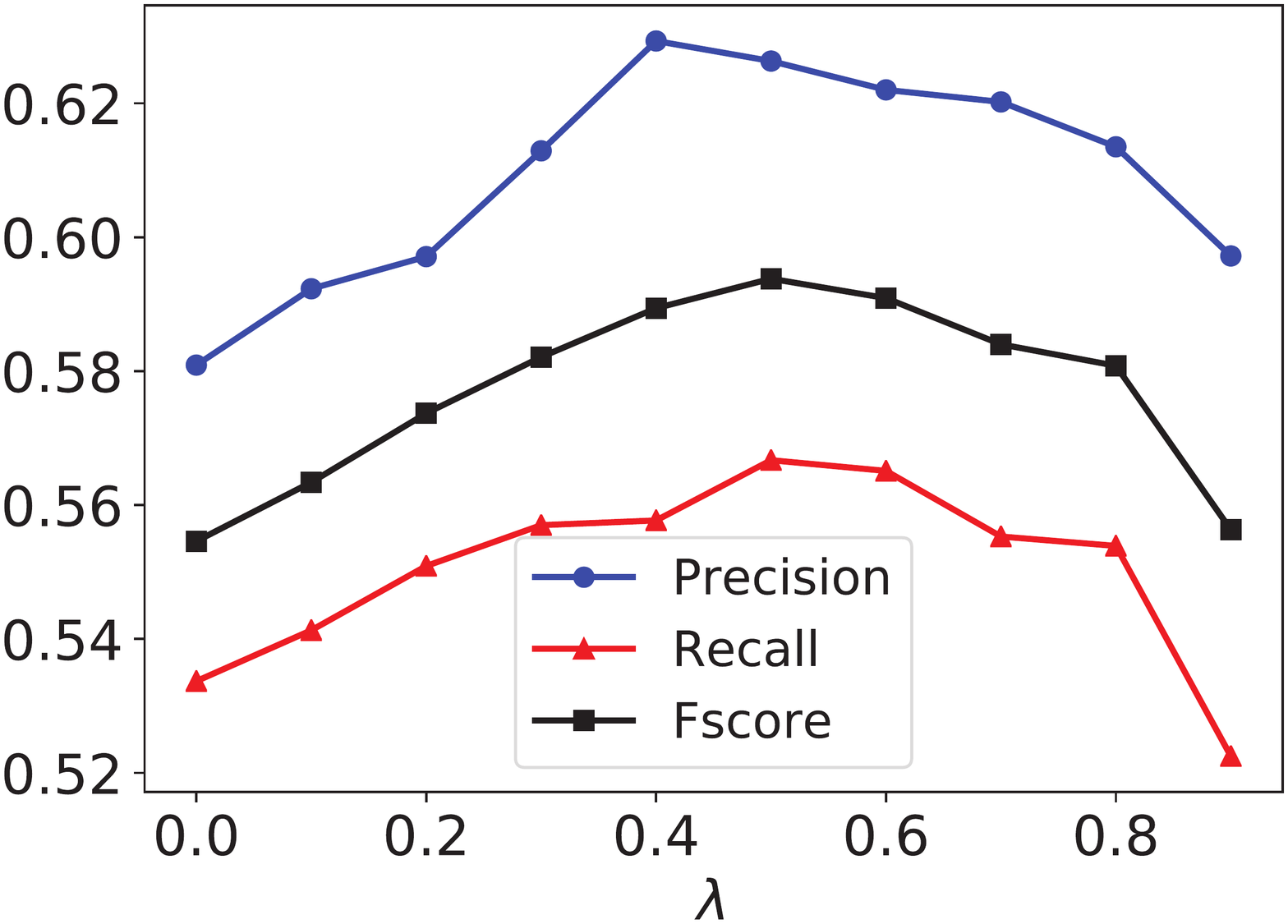}
	\caption{The influence of $\lambda$ value.}\label{fig.lambda}
\end{figure}

In this section we conduct experiments to explore the influence of hyperparameters on the performance of our approach.
Due to space limit, here we only show the influence of the most important hyperparameter on our approach, i.e., $\lambda$ in Eq.~(\ref{eq.loss_all}).
The results are summarized in Figure~\ref{fig.lambda}.
This experiment was conducted on the \textit{Bakeoff-4} dataset with 5\% training data.
$\lambda$ is used to control the relative importance of CWS loss in the overall loss of our joint training framework (Eq.~(\ref{eq.loss_all})).
According to Figure~\ref{fig.lambda}, the performance of our approach improves when $\lambda$ increases from 0, and declines when $\lambda$ becomes too large.
This is because when $\lambda$ is too small, the useful information in word segmentation is not fully exploited. 
Thus, the performance is not optimal.
When $\lambda$ becomes too large, the auxiliary CWS task is over-emphasized and the loss of NER is not fully respected.
Thus, the performance is also not optimal.
A moderate value of $\lambda$ is most suitable for our approach.

\section{Conclusion}\label{sec:Conclusion}

In this paper we propose a neural approach for Chinese named entity recognition.
First, we propose a CNN-LSTM-CRF neural architecture for Chinese NER, where CNN and Bi-LSTM networks are used to learn character representations from both local and long-distance contexts, and CRF is used for joint label decoding.
Second, we propose a unified framework to jointly train Chinese named entity recognition and word segmentation models to exploit the inherent relatedness between these two tasks and enhance the ability of Chinese NER model in identifying entity boundaries.
In addition, we propose an automatic method to generate pseudo labeled samples from existing labeled data to enrich the training data.
Extensive experiments on two benchmark datasets validate that our approach can effectively improve the performance of Chinese named entity recognition especially when training data is insufficient, and can consistently outperform many baseline methods.

\begin{acks}
This work was supported by the National Key Research and Development Program of China under
Grant number 2018YFC1604002, and the National Natural Science Foundation of China under Grant numbers U1705261, U1536201, and U1536207.
\end{acks}

\bibliographystyle{ACM-Reference-Format}
\balance 
\bibliography{CNER}

\end{CJK*} 
\end{document}